\documentclass[runningheads]{llncs}

\usepackage{accv}

\usepackage{accvabbrv}

\usepackage{graphicx}
\usepackage{booktabs}
\usepackage{multirow}
\usepackage{tabularx}
\usepackage{array}
\usepackage{placeins}
\usepackage{amsmath}

\usepackage[accsupp]{axessibility}  

\usepackage[pagebackref,breaklinks,colorlinks,citecolor=accvblue]{hyperref}
\usepackage{hyperref}

\usepackage{orcidlink}

\begin{document}

\title{DEQ-Stereo: Deep Equilibrium Video Stereo Matching via Adaptive Convergence}

\titlerunning{DEQ-Stereo}

\author{Kiljoon Han\inst{1}\orcidlink{0009-0001-7276-8858} \and
Kyumin Hwang\inst{2}\orcidlink{0009-0000-7878-697X} \and
Wonhyeok Choi\inst{1}\orcidlink{0009-0005-3270-8083} \and
Jeonghoon Kim\inst{1}\orcidlink{0000-0002-7568-4115} \and
Sunghoon Im\inst{2}$^{,\mathsf{\dagger}}$\orcidlink{0000-0001-9776-8101}}

\authorrunning{K.~Han et al.}

\institute{Daegu Gyeongbuk Institute of Science and Technology, Daegu, South Korea \and
Korea Advanced Institute of Science and Technology, Daejeon, South Korea}

\maketitle

\begin{abstract}
Iterative refinement models have significantly improved stereo matching accuracy, but their rigid, fixed update schedules incur heavy computational costs and a strict trade-off between speed and accuracy. This challenge is magnified in video streams, where traditional online methods resort to reducing feature channels or relying on expensive temporal warping. To address this, we propose \textbf{DEQ-Stereo}, a novel video stereo matching framework based on the Deep Equilibrium (DEQ) paradigm. Instead of unrolling a fixed number of iterations, DEQ-Stereo reformulates the refinement as an implicit fixed-point problem, adaptively terminating once convergence is met. To ensure fast and stable intra-frame convergence, we introduce the \textbf{Convolutional Simplified Gated Unit (ConvSGU)}, which eliminates unnecessary non-linearities by modeling the latent state as a gated, element-wise sum of inputs. Furthermore, to exploit cross-frame temporal coherence, we propose \textbf{Temporal State Transfer (TST)}, a warm-start strategy that propagates the converged hidden state to the next frame without requiring explicit pose or warping priors.
Extensive evaluations demonstrate that DEQ-Stereo delivers state-of-the-art performance with a superior accuracy-to-latency trade-off.

\keywords{Stereo Matching \and Deep Equilibrium Models \and Video Depth}
\end{abstract}
\section{Introduction}

\begin{figure}[htb]
     \centering
     \includegraphics[width=0.85\linewidth]{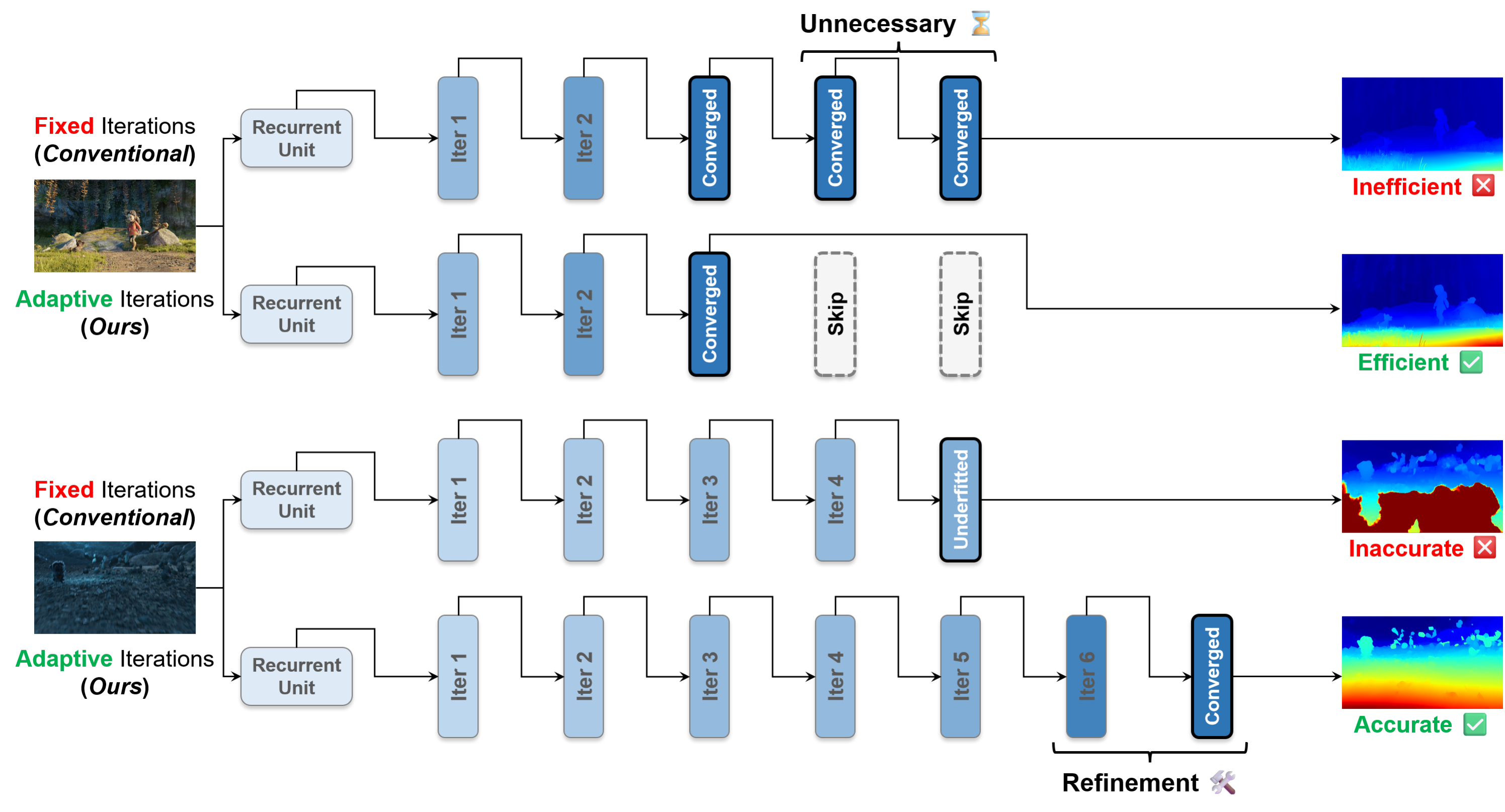}
     \caption{\textbf{Comparisons of Iterative Stereo Matching.} Traditional methods (e.g., IGEV-Stereo) use a fixed number of iterations, resulting in a trade-off between accuracy and efficiency. In contrast, our method adaptively determines the number of iterations, achieving high accuracy and efficiency.}
     \label{fig:motivation}
\end{figure}

Stereo matching is a fundamental problem in computer vision, aimed at estimating depth by computing disparities between a pair of images. It underpins various 3D perception tasks, including autonomous driving~\cite{kitti2012, kitti2015}, robotics~\cite{zhang2015building}, augmented reality~\cite{cheng2024stereo}, and 3D scene reconstruction~\cite{cabon2025must3r}.
Recent learning-based approaches, particularly those based on iterative refinement~\cite{lipson2021raft,xu2023iterative,li2022practical}, have significantly improved accuracy. These methods refine the disparity map across multiple steps, progressively enlarging the receptive field and enabling robust predictions in challenging areas like textureless regions~\cite{zhao2023high, li2022practical} or thin structures~\cite{wang2024selective, chen2024mocha}. Their generalization ability and optimization-like behavior have made them widely adopted in both static and sequential stereo settings. However, their inference time increases linearly with the number of iterations, which can be a bottleneck in time-sensitive environments.

While stereo matching is often benchmarked on individual image pairs, many real-world applications require continuous stereo video processing. Applying iterative models independently to each frame leads to redundant computation and temporal inconsistencies across frames.
Recent video stereo methods incorporate temporal cues to improve both runtime and coherence.
These can be broadly categorized into offline and online approaches. Offline methods~\cite{karaev2023dynamicstereo, jing2024match} exploit full sequences for improved accuracy but are impractical for real-time use, as they rely on future frames. Online methods~\cite{zhang2023temporalstereo, zeng2024temporally, li2023temporally} are designed for real-time processing by using lightweight models and previous-frame information, but suffer from accumulated errors and require reliable pose and disparity priors.

To address the limitations of existing video stereo matching methods, we propose DEQ-Stereo, a novel framework based on the Deep Equilibrium (DEQ) model~\cite{bai2019deep}, designed for efficient and temporally consistent stereo matching in video sequences. Unlike conventional approaches that rely on a fixed number of iterative updates, DEQ-Stereo directly solves for the fixed point of the update module, eliminating the need for explicit unrolling and enabling adaptive inference based on a convergence criterion as shown in \cref{fig:motivation}. This yields more efficient, stable, and scalable inference across frames.
To support this equilibrium-based formulation, we propose two complementary strategies aimed at improving intra-frame convergence stability and enhancing robustness across frames.

First, we introduce the Convolutional Simplified Gated Unit (ConvSGU), designed to accelerate and stabilize per-frame convergence.
Inspired by Recurrent Additive Networks (RAN)~\cite{lee2017recurrentadditivenetworks}, ConvSGU simplifies the gating structure of ConvGRU by expressing the latent state as a gated, element-wise sum of inputs, thereby removing unnecessary non-linearities.
This simplification facilitates faster and more stable convergence to equilibrium states within the DEQ formulation.
Furthermore, to exploit temporal coherence in video, we propose Temporal State Transfer (TST), a mechanism that reuses the converged hidden state from the previous frame to initialize the current one. This warm-start strategy encourages consistency over time and improves convergence speed, particularly in scenes with low inter-frame motion. By integrating the DEQ paradigm with ConvSGU and TST, DEQ-Stereo produces temporally consistent, high-quality disparity estimates while significantly reducing computational cost. Extensive evaluation across multiple benchmarks including SceneFlow~\cite{sceneflow}, TartanAir~\cite{wang2020tartanair}, KITTI2015~\cite{kitti2015}, and ETH3D SLAM~\cite{schops2019bad}, demonstrates that our method achieves the best trade-off between efficiency and accuracy.

Our main contributions can be summarized as follows:
\begin{enumerate}
    \item We propose \textbf{DEQ-Stereo}, a novel stereo matching framework leveraging Deep Equilibrium models for adaptive and accurate iterative refinement.
    \item To improve convergence from both intra-frame and inter-frame perspectives, we propose \textbf{ConvSGU} enhancing convergence stability within each frame, and \textbf{TST} accelerating convergence across video frames.
    \item Our method achieves state-of-the-art performance across multiple online video stereo benchmarks, delivering the best trade-off between inference speed and accuracy.
\end{enumerate}
\section{Related Work}
\subsection{Per-frame Stereo Matching Networks}
Early deep learning stereo methods centered on building a cost volume and regularizing it with computationally expensive 3D convolutions to directly regress disparity~\cite{chang2018pyramid, kendall2017end, guo2019group, zhang2019ga}.
While these methods laid the groundwork, their high computational cost motivated further refinements in cost aggregation strategies~\cite{xu2022attention, liu2022local}.
A significant shift occurred with the introduction of iterative refinement architectures, which adapted RAFT~\cite{teed2020raft} from optical flow for stereo matching~\cite{lipson2021raft}.
This design, while powerful, relies on a fixed number of iterations—a core limitation forcing a trade-off between accuracy and efficiency.
Subsequent work has focused on improving performance within this fixed iteration, either by enhancing frameworks with hierarchical processing or better geometric representations~\cite{li2022practical, xu2023iterative, zeng2023parameterized}, or by designing specialized recurrent modules to preserve high-frequency details~\cite{zhao2023high, wang2024selective, chen2024mocha}.
However, despite these advances, the fundamental issue of fixed iteration remains.

\subsection{Video Stereo Matching Networks}
In video stereo matching, methods leverage temporal information for efficiency and consistency.
While offline approaches~\cite{karaev2023dynamicstereo, jing2024match, jing2025stereo} yield high-quality, temporally smooth results, their non-causal nature makes them unsuitable for real-time applications.
Online methods~\cite{li2023temporally, zeng2024temporally, cheng2024stereo} are therefore more common, ranging from early RNN‑based designs~\cite{zhong2018open} to lightweight temporal modules and adaptations of iterative refinement architectures prominent in per‑frame matching.
Focusing on the latter, these iterative approaches operate with a low, fixed number of iterations, supplementing their limited refinement capabilities by warping the depth from the previous frame to provide a strong initial estimate.
This warping strategy, however, necessitates auxiliary camera parameters, and its direct injection of prior estimates hinders the model's ability to correct errors, leading to their accumulation over time.
This specific vulnerability of fixed, low-iteration schemes directly motivates our adaptive approach.

\subsection{Implicit Deep Learning}
Unlike traditional fixed-sequence layers, Implicit Deep Learning networks—such as Deep Equilibrium Models (DEQ)~\cite{bai2019deep} and Neural ODEs~\cite{chen2018neural}—define outputs as fixed-point solutions of a parameterized process, enabling effectively infinite depth. By adaptively determining the iterations needed to reach equilibrium, DEQ overcomes the inefficiencies of fixed-iteration stereo matching, yielding significant computational savings without sacrificing refinement quality. This implicit paradigm has been successfully applied across diverse domains, including diffusion models~\cite{pokle2022deep, cao2024deep}, object detection~\cite{wang2023deep}, and graph networks~\cite{gu2020implicit}.
In dense correspondence, DEQ-Flow~\cite{bai2022deep} pairs the DEQ framework with ConvGRU for optical flow, reusing both the predicted flow and hidden state from the previous frame to accelerate convergence. While our TST shares this warm-start intuition, it differs crucially by propagating only the hidden state. Empirically, we find that this selective propagation reduces sensitivity to complex scene dynamics, ultimately improving temporal stability in video stereo matching.
\section{Preliminaries}
\label{sec:pre}

\subsection{Iterative Stereo Matching}
Iterative stereo matching methods~\cite{lipson2021raft,li2022practical,xu2023iterative} progressively refine disparity estimates. This process can be defined as a parameterized function $f_\theta$ that updates hidden state $\mathbf{h}^k$ and disparity $\mathbf{d}^k$ at iteration $k$ using the correlation volume $\mathbf{C}$:
\begin{equation}
    [\mathbf{h}^{k+1}, \mathbf{d}^{k+1}] = f_\theta(\mathbf{h}^k, \mathbf{d}^k, \mathbf{C}).
\label{equ:f_theta}
\end{equation}
Internally, $f_\theta$ typically employs a recurrent unit (e.g., ConvGRU) represented as $\mathcal{R}_{\theta}$, operating as follows:
\begin{equation}
\begin{gathered}
    \mathbf{x}^k=\text{MotionEnc}(\mathbf{d}^{k}, \text{LookUp}(\mathbf{C},\mathbf{d}^{k} )),\quad
    \mathbf{h}^{k+1} = \mathcal{R}_\theta(\mathbf{h}^{k},\mathbf{x}^k), \\
    \mathbf{d}^{k+1} = \mathbf{d}^{k} + \mathrm{DispHead}(\mathbf{h}^{k+1}),
\end{gathered}
\label{equ:rnn}
\end{equation}
where $\mathbf{x}^k$ is the motion feature extracted by $\mathrm{MotionEnc}(\cdot, \cdot)$ consisting of two convolutional layers, $\mathrm{LookUp}(\cdot, \cdot)$ samples local geometry from the correlation volume $\mathbf{C}$ guided by $\mathbf{d}^k$ as proposed in previous works \cite{teed2020raft, lipson2021raft}, and $\mathbf{d}^{k+1}$ is updated by adding the residual disparity predicted by the disparity head $\text{DispHead}(\cdot)$.

Iterative methods excel by aggregating features across multiple recursive steps, effectively combining local matching cues with global context.
A key limitation is their reliance on a fixed iteration count $N$, chosen empirically rather than based on explicit convergence criteria.
This results in a trade-off between accuracy and efficiency: insufficient iterations impair disparity quality, while excessive iterations increase inference time.
This balance becomes even more critical in video stereo settings, where real-time constraints and dynamic scene content demand adaptive iteration control to optimize both speed and accuracy.

\subsection{Deep Equilibrium Paradigm}
 The DEQ~\cite{bai2019deep,bai2022deep} reformulates the update operation to an implicit function and finds a joint equilibrium state $\mathbf{s}^*=[\mathbf{h}^*,\mathbf{d}^*]$ that satisfies:
\begin{equation}
    \begin{gathered}
    \mathbf{s}^*=[\mathbf{h}^*,\mathbf{d}^*] = f_\theta(\mathbf{h}^*,\mathbf{d}^*, \mathbf{C}) \quad
    \Leftrightarrow \quad
    g_\theta(\mathbf{s}^*, \mathbf{C}) \equiv f_\theta(\mathbf{s}^*, \mathbf{C}) - \mathbf{s}^* \rightarrow 0,
    \end{gathered}
    \label{equ:fixed_point_prelim}
\end{equation}
where $\mathbf{s}^*$ is the joint state that no longer changes under further applications of an update function $f_\theta$, and $g_\theta$ is termed the residual function.
The core idea of DEQ is to find a joint equilibrium state $\mathbf{s}^*$ by finding the root of $g_\theta(\mathbf{s}^*, \mathbf{C}) = 0$ through a numerical solver, instead of performing a fixed number of explicit iterations.
This numerical solver performs the forward pass by iteratively updating the joint state $\mathbf{s}$.
At each solver step $k$, the state $\mathbf{s}^k$ is updated to $\mathbf{s}^{k+1}$ as follows:
\begin{equation}
    \begin{gathered}
        \mathbf{s}^{k+1} = \mathbf{s}^k - \alpha \mathbf{J}^{-1}_{g_\theta | \mathbf{s}^{k}} g_\theta(\mathbf{s}^k, \mathbf{C}), \quad \mathbf{J}^{-1}_{g_\theta | \mathbf{s}^{k}} \approx \mathbf{B}^{k}_{g_\theta}, \\[0.3em]
        \mathbf{B}^{k}_{g_\theta}= \mathbf{B}^{k-1}_{g_\theta} + \frac{\Delta \mathbf{s}^{k} - \mathbf{B}^{k-1}_{g_\theta}\Delta g^{k}_\theta}{\Delta {\mathbf{s}^{k}}^{\top} \mathbf{B}^{k-1}_{g_\theta}\Delta g^{k}_\theta} \Delta {\mathbf{s}^{k}}^{\top} \mathbf{B}^{k-1}_{g_\theta},
    \end{gathered}
    \label{equ:deq_forward}
\end{equation}
where $\alpha$ denotes the step size, and $\mathbf{J}^{-1}_{g_\theta | \mathbf{s}^{k}}$ is the inverse Jacobian of $g_\theta$ with respect to the current state $\mathbf{s}^k$, which is essential for determining the update direction towards the root of $g_\theta$.
The update rule for $\mathbf{B}_{g_\theta}^{k}$ (\cref{equ:deq_forward}) is Broyden's method, a quasi-Newton approach that efficiently approximates the inverse Jacobian to determine the update direction.
While this method provides the direction, the step size $\alpha$ is crucial for ensuring stable convergence.
For this purpose, an Armijo backtracking line search~\cite{armijo1966minimization} is employed to adaptively find a step size that guarantees a sufficient decrease in the norm of the residual function, $||g_\theta(\mathbf{s}^*, \mathbf{C})||$, ensuring robust progress towards the joint equilibrium state.
\section{Method}

\begin{figure*}[h!]
     \centering
     \includegraphics[width=1.0\linewidth]{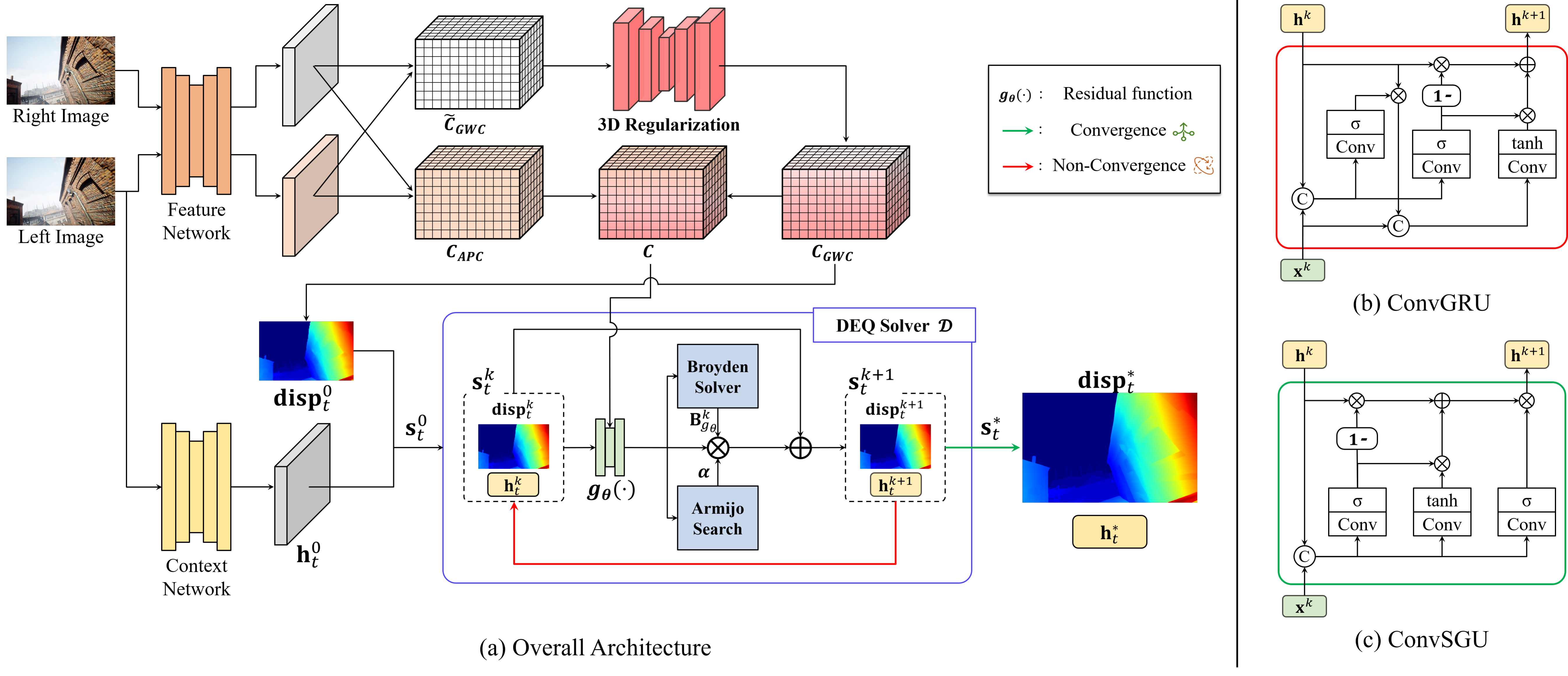}
     \caption{\textbf{Overall pipeline.} (a) Overview of DEQ-Stereo. From stereo pairs, we construct a correlation volume and regress an initial disparity, which is refined via an implicit, convergence-driven DEQ process. Unlike prior iterative methods, DEQ-Stereo supports adaptive termination and faster convergence, enabled by two key components: ConvSGU and Temporal State Transfer. We denote the frame index as $t$, the DEQ iteration step as $k$, and the converged state as $*$. \textbf{(b)} Standard ConvGRU structure. \textbf{(c)} Our proposed ConvSGU, a simplified variant of ConvGRU, which defines the residual function $g_\theta(\cdot)$ for the DEQ solver.}
     \label{fig:overall_arch}
\end{figure*}
\subsection{Deep Equilibrium Stereo Matching}
Building upon the principles of Deep Equilibrium (DEQ) outlined in \cref{sec:pre}, particularly their capacity for adaptive termination, we propose DEQ-Stereo as illustrated in \cref{fig:overall_arch}.
This approach replaces the standard recurrent refinement module in iterative stereo matching with a DEQ module to overcome the inefficiencies of fixed-iteration schemes.
DEQ-stereo obtains the left and right images $\mathbf{I}_L$ and $\mathbf{I}_R$ as input, designed to produce the disparity map $\mathbf{d}$ between two images. 
We first extract features $\mathbf{f}_L, \mathbf{f}_R$ of images using the shared feature encoder $\mathrm{FeatureNet}(\cdot)$. 
In parallel, the left image is passed through a context network $\mathrm{ContextNet}(\cdot)$ to generate an initial hidden state $\mathbf{h}^0$ as follows:
\begin{equation}
\begin{gathered}
\mathbf{f}_{L} = \mathrm{FeatureNet}(\mathbf{I}_L), \quad\mathbf{f}_{R} = \mathrm{FeatureNet}(\mathbf{I}_R), \quad
\mathbf{h}^0 = \mathrm{ContextNet}(\mathbf{I}_L).
\end{gathered}
\end{equation}
Following IGEV-Stereo~\cite{xu2023iterative}, we then construct a correlation volume $\mathbf{C} \in $ $\mathbb{R}^{(N_g+1) \times D\times H\times W}$ composed of one All-Pairs Correlation $\mathbf{C}_{APC}$ and $N_g$ Group-Wise Correlations $\mathbf{C}_{GWC}$. From the $\mathbf{C}_{GWC}$, we then regress an initial disparity map $\mathbf{d}^0$. The volume and initial disparity regression are formulated as follows:
\begin{equation}
\begin{gathered}
    \mathbf{d}^0 = \sum^{D-1}_{d=0} d\times \text{Softmax}(\mathbf{C}_{GWC}(d)),~~\mathbf{C}_{GWC} = \mathrm{Reg}\big(\frac{N_g}{N_c}\langle \mathbf{f}_L^g(x,y), \mathbf{f}_R^g(x - d, y) \rangle\big), 
    \\
    \mathbf{C} = [\mathbf{C}_{APC}, \mathbf{C}_{GWC}],~~ \mathbf{C}_{APC}(d, x, y) = \langle \mathbf{f}_L(x,y), \mathbf{f}_R(x - d, y) \rangle,
\end{gathered}
\end{equation}
where $d$ is the disparity index, $g$ is the group index for $\mathbf{C}_{GWC}$, $N_c$ is the number of feature channels, $N_g$ is the number of groups, $\mathrm{Reg}(\cdot)$ denotes the regularization network, and $[\cdot, \cdot]$ denotes channel-wise concatenation.

With the initial hidden state $\mathbf{h}^0$ and disparity $\mathbf{d}^0$, we define the joint state as $\mathbf{s}^k = [\mathbf{h}^k, \mathbf{d}^k]$. At each solver step $k \in \{0,...,K_{\text{stop}}\}$,
the DEQ solver $\mathcal{D}$ iteratively updates the current joint state $\mathbf{s}^k$ using the correlation volume $\mathbf{C}$ and the residual function $g_{\theta}$, following the update rule defined in \cref{equ:deq_forward}.
The DEQ solver $\mathcal{D}$ iteratively updates the joint state, terminating after $K_{\text{stop}}$ steps when the following stopping criterion is satisfied:
\begin{equation}
\label{equ:criteria}
\| g_\theta(\mathbf{s}^k, \mathbf{C}) \| < \epsilon \quad \text{or} \quad k \geq \text{MaxIter},
\end{equation}
where $\epsilon$ is a convergence threshold, and $\text{MaxIter}$ is the iteration limit.
The actual number of steps until convergence is recorded as $K_{\text{stop}}$.

\subsection{Convolutional Simplified Gated Unit}
In DEQ-based networks, the convergence speed of the residual function critically affects inference efficiency. While ConvGRU provides expressive modeling through gating, its complex nonlinear update rules can impede convergence. To address this, we propose the Convolutional Simplified Gated Unit (ConvSGU), a streamlined variant of ConvGRU inspired by the Recurrent Additive Network (RAN)~\cite{lee2017recurrentadditivenetworks}. ConvSGU simplifies recurrent dynamics while retaining modeling capacity, defined as:
\begin{equation}
\label{equ:ConvSGU}
\begin{gathered}
\mathbf{z} = \sigma(\text{Conv}([\mathbf{h}^k, \mathbf{x}^k], W_i) + \mathbf{b}_i), \quad
\mathbf{f} = \sigma(\text{Conv}([\mathbf{h}^k, \mathbf{x}^k], W_f) + \mathbf{b}_f), \\
\mathbf{c} = \tanh(\text{Conv}([\mathbf{h}^k, \mathbf{x}^k], W_c) + \mathbf{b}_c), \quad
\mathbf{h}^{k+1} = (\mathbf{z}\cdot \mathbf{c} +(1-\mathbf{z})\cdot \mathbf{h}^k)\cdot \mathbf{f},
\end{gathered}
\end{equation}
where $\sigma$ is the activation function, and $\mathbf{b}_i$, $\mathbf{b}_f$, and $\mathbf{b}_c$ are constant inputs generated from the left image.
The variables $\mathbf{z}$, $\mathbf{f}$, and $\mathbf{c}$ denote the update gate, forget gate, and candidate cell state, respectively.
ConvSGU modifies the standard ConvGRU by removing the reset gate—eliminating unstable multiplicative dynamics~\cite{lee2017recurrentadditivenetworks, bai2019deep}—and introducing a forget gate $\mathbf{f}$ as a dynamic damping factor. These changes smooth the residual function and enhance convergence stability in the DEQ framework.

\begin{figure}[t]
     \centering
     \includegraphics[width=0.9\linewidth]{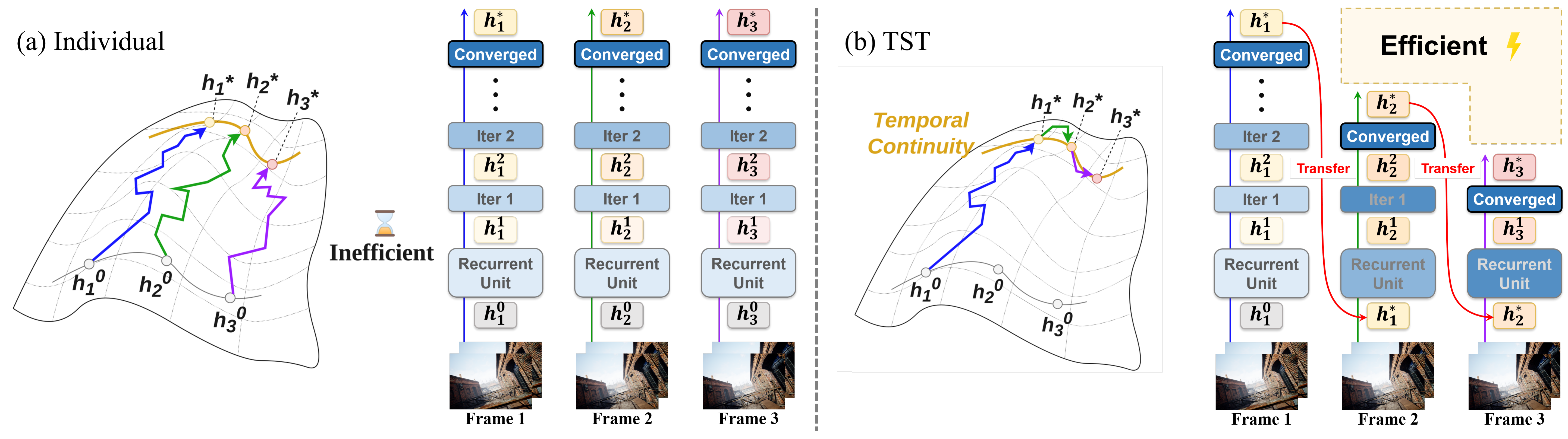}
     \caption{\textbf{Illustration of TST.} The output state from the previous frame is reused to initialize the current frame, allowing faster and more stable disparity refinement.}
     \label{fig:tst}
\end{figure}

\subsection{Temporal State Transfer}
\label{sec:tst}

For sequential inputs such as video data, consecutive frames exhibit strong temporal coherence.
We introduce Temporal State Transfer (TST) to exploit this property, a mechanism designed exclusively to accelerate the inference of DEQ-Stereo in video settings.
Inspired by Fixed-point Reuse in DEQ-Flow~\cite{bai2022deep}, TST selectively reuses the converged hidden state $\mathbf{h}_{t-1}^*$ from the previous joint equilibrium state $\mathbf{s}_{t-1}^* = [\mathbf{h}_{t-1}^*, \mathbf{d}_{t-1}^*]$.
The converged disparity $\mathbf{d}_{t-1}^*$, however, is not reused; instead, the initial disparity $\mathbf{d}_t^0$ is freshly regressed from the current correlation volume $\mathbf{C}_t$.
By combining these, we construct the initial joint state for the current frame $t \in \{1,...,T\}$.
Formally, the DEQ solver $\mathcal{D}$ for the sequence is defined as follows:
\begin{equation}
\label{equ:tst}
    \mathbf{s}_t^* = {\mathcal{D}}(\mathbf{s}_t^0, g_{\theta}, \mathbf{C}_t) = {\mathcal{D}}([\mathbf{h}_{t-1}^*, \mathbf{d}_t^0], g_{\theta}, \mathbf{C}_t).
\end{equation}

The intuition is that the previous hidden state $\mathbf{h}_{t-1}^*$ serves as a strong initial guess for frame $t$, especially when inter-frame changes are small.
This approach significantly reduces the number of iterations needed per frame, resulting in much faster inference speed that is highly advantageous for real-time video stereo applications.
By leveraging temporal continuity, TST also enhances temporal consistency across frames in most scenarios.

\subsection{Loss Function}
For training, gradients $\frac{\partial\mathcal {L}}{\partial\theta}$ are efficiently computed using the Implicit Function Theorem (IFT)~\cite{krantz2002implicit,bai2019deep,bai2022deep}.
\begin{equation}
\label{equ:ift_grad}
\begin{gathered}
    {\frac{\partial \mathcal{L}}{\partial \theta}} = {\frac{\partial \mathcal{L}}{\partial \mathbf{s}^*}}{\frac{\mathrm{d} \mathbf{s}^*}{\mathrm{d} \theta}} 
    = -{\frac{\partial \mathcal{L}}{\partial \mathbf{s}^*}}(\mathbf{J}^{-1}_{g_{\theta}|\mathbf{s}^*}){\frac{\partial f_\theta(\mathbf{s}^*, \mathbf{C})}{\partial \theta}},
\end{gathered}
\end{equation}
where $\frac{\partial\mathcal{L}}{\partial \mathbf{s}^*}$ represents the gradient of the loss with respect to the joint equilibrium state.
A significant advantage of using IFT is that it enables backpropagation using only the converged state $\mathbf{s}^*$, without needing to store intermediate solver iterates.
However, since the DEQ module does not consist of discrete, explicitly unrolled layers, applying loss solely to the final converged state $\mathbf{s}^*$ might not sufficiently supervise the model to ensure convergence to a desirable equilibrium.

To address this and better guide the learning process towards meaningful equilibrium, inspired by DEQ-Flow~\cite{bai2022deep}, we introduce an auxiliary loss term computed on randomly selected intermediate disparity estimates. Our loss term is defined as follows:
\begin{equation}
\label{equ:loss}
\begin{aligned}
    \mathcal{L}_{\mathrm{total}} =\mathcal{L}_{\mathrm{eq}} + \mathcal{L}_{\mathrm{int}} =||\mathbf{d}^*-\mathbf{d}_{gt}||^2_2 + \gamma||\mathbf{d}^{[i]}-\mathbf{d}_{gt}||^2_2,
\end{aligned}
\end{equation}
where $\mathbf{d}_{gt}$ is the ground truth disparity, $\mathbf{d}^{*}$ is the converged disparity, $\mathbf{d}^{[i]}$ is the disparity predicted from a randomly chosen intermediate hidden state during the solver's iterations, and $\gamma <1$ is a hyperparameter weighting the intermediate loss. This combined loss encourages both the final converged state to be accurate and the iterative refinement process itself to progress towards useful solutions.

\section{Experiments}

\subsection{Implementation Details}

\noindent \textbf{Training Setup.}
We implement DEQ-Stereo in PyTorch, training on eight NVIDIA TITAN RTX GPUs and evaluating on a single RTX 3090.
We use AdamW~\cite{loshchilov2017decoupled} with gradients clipped to $[-1,1]$.
For pretraining, we separately train on SceneFlow~\cite{sceneflow} and the TartanAir TemporalStereo split~\cite{wang2020tartanair,zhang2023temporalstereo} for 200K steps with a one-cycle learning rate schedule with a learning rate of $2\times10^{-4}$ and a total batch size of 8.
For KITTI fine-tuning, we initialize from the TartanAir model and fine-tune on KITTI raw~\cite{Geiger2013IJRR} using LEAStereo~\cite{cheng2020hierarchical} pseudo-labels, following TC-Stereo~\cite{zeng2024temporally}.
We train for 50K steps with a maximum learning rate of $1\times10^{-4}$.
The TartanAir model is directly used for zero-shot evaluation on ETH3D SLAM~\cite{schops2019bad}.
All training is performed on a per-frame basis, while the video mode during inference is determined by whether TST is enabled.

\noindent \textbf{DEQ Solver \& Inference.}
The DEQ solver uses Broyden's method with Armijo backtracking for stable convergence, with an Armijo constant of $1\times10^{-4}$, initial step size of $0.5$, and minimum step size of $1\times10^{-2}$.
It terminates when $\mathrm{max}(||g_\theta(\mathbf{s},\mathbf{C})||)$ falls below $1\times10^{-3}$ during training or 30 during inference, or after 16 iterations.
Without TST, each frame is initialized from scratch; with TST, the converged state is propagated to the next frame to exploit temporal coherence.
All other solver settings remain identical between the two modes.

\begin{figure*}[t!]
    \centering
    \begin{minipage}[t]{0.585\linewidth}
        \centering
        \captionof{table}{Quantitative results on TartanAir. The \textbf{best} and \underline{second-best} are bolded and underlined.}
        \resizebox{0.94\textwidth}{!}{
\begin{tabular}{c|ccc|ccc}
        \toprule[1.5pt]
        \multirow{3}{*}{Method}    & \multicolumn{3}{c|}{OCC}                                    & \multicolumn{3}{c}{ALL}              \\ \cmidrule{2-7}
                                   & EPE               & Bad 3.0            & Bad 1.0            & EPE               & Bad 3.0           & Bad 1.0   \\
                                   & (px)              & (\%)               & (\%)               & (px)              & (\%)              & (\%)      \\ \midrule
        CODD                       & -                 & -                  & -                  & 0.90              & 5.30              & -         \\
        TemporalStereo             & 1.60              & 18.46              & 39.76              & 0.66              & 3.81              & 11.18     \\
        TC-Stereo                  & \underline{1.36}  & \underline{16.36}  & \underline{34.68}  & \underline{0.52}  & \underline{3.31}  & \underline{8.51} \\ \midrule
        DEQ-Stereo (Ours)          & \textbf{0.77}     & \textbf{4.48}      & \textbf{11.20}     & \textbf{0.39}     & \textbf{1.88}     & \textbf{5.12}      \\
        \bottomrule[1.5pt]
\end{tabular}
}
        \label{tab:tartan}
    \end{minipage}
    \hfill
    \begin{minipage}[t]{0.375\linewidth}
        \centering
        \captionof{table}{Quantitative results on SceneFlow.}
        \resizebox{0.88\textwidth}{!}{
\begin{tabular}{l|cc}
    \toprule[1.5pt]
    \multirow{2}{*}{Method}   & EPE               & Bad 3.0           \\
                              & (px)              & (\%)              \\ \midrule
    CODD                      & 0.60              & 2.80              \\
    TemporalStereo            & 0.53              & 2.75              \\
    TC-Stereo                 & \underline{0.50}  & \underline{2.39}  \\ \midrule
    DEQ-Stereo (Ours)         & \textbf{0.45}     & \textbf{2.37}     \\
    \bottomrule[1.5pt]
\end{tabular}
}
        \label{tab:sceneflow}
    \end{minipage}
\end{figure*}

\begin{figure*}[t!]
     \centering
     \includegraphics[width=0.99\linewidth]{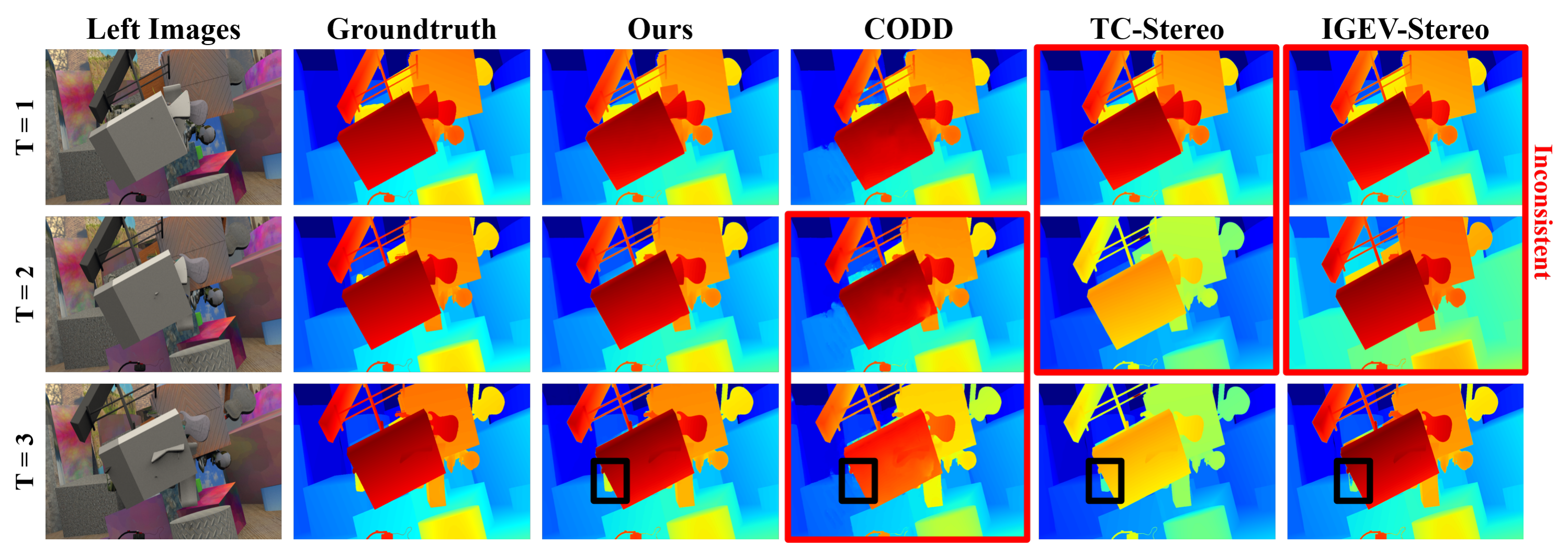}
     \caption{\textbf{Visualizations of disparity results on SceneFlow dataset}. Compared to CODD, TC-Stereo (video-based) and IGEV-Stereo (per-frame), our method shows more temporally consistent disparity predictions ({\color{red}\textbf{red box}}) and better preserves object boundaries in ill-posed regions where temporal cues are unavailable (\textbf{black box}).}
     \label{fig:qual}
\end{figure*}
\begin{table}[t!]
\centering
\caption{Quantitative results on KITTI2015. A method$^\dagger$ requires camera parameters as auxiliary input. Note that the TST cannot be applied to per-frame methods.}
\resizebox{0.85\textwidth}{!}{
\setlength{\tabcolsep}{8pt}
\begin{tabular}{c|c|ccc|ccc|c}
\toprule[1.5pt]
\multirow{3}{*}{Type} & \multirow{3}{*}{Method}
                    & \multicolumn{3}{c|}{NOC}                    & \multicolumn{3}{c|}{ALL}                & \multirow{3}{*}{\shortstack{Time\\(s)}} \\ \cmidrule{3-8}
                    && D1-bg             & D1-fg        & D1-all       & D1-bg         & D1-fg         & D1-all  &                   \\
                    && (\%)              & (\%)         & (\%)         & (\%)          & (\%)          & (\%)    &                   \\ \midrule
\multirow{6}{*}{\shortstack{Per\\frame}}
& RAFT-Stereo               & 1.45 & 2.94 & 1.69 & 1.58 & 3.05 & 1.82 & \underline{0.30} \\
& IGEV-Stereo               & 1.27 & 2.62 & 1.53 & 1.38 & 2.67 & 1.59 & 0.34 \\
& Selective-IGEV            & \underline{1.22} & 2.55 & \underline{1.44} & \underline{1.33} & 2.61 & 1.55 & 0.45 \\
& MoCha-Stereo              & 1.24 & \textbf{2.42} & \underline{1.44} & 1.36 & \textbf{2.43} & \underline{1.53} & 0.50 \\ \cmidrule{2-9}
& DEQ-Stereo & \multirow{2}{*}{\textbf{1.20}} & \multirow{2}{*}{\underline{2.47}} & \multirow{2}{*}{\textbf{1.41}} & \multirow{2}{*}{\textbf{1.31}} & \multirow{2}{*}{\underline{2.49}} & \multirow{2}{*}{\textbf{1.51}} & \multirow{2}{*}{\textbf{0.16}} \\
& (Ours) w/o TST &&&&&&& \\ \midrule
\multirow{4}{*}{Video}
& TemporalStereo$^{\dag}$        & 1.52 & 2.58 & 1.70 & 1.61 & 2.78 & 1.81 & \textbf{0.05}  \\
& TC-Stereo$^{\dag}$             & \textbf{1.21} & \textbf{2.24} & \textbf{1.38} & \textbf{1.29} & \textbf{2.33} & \textbf{1.46} & 0.11 \\ \cmidrule{2-9}
& DEQ-Stereo  & \multirow{2}{*}{\textbf{1.21}} & \multirow{2}{*}{\underline{2.52}} & \multirow{2}{*}{\underline{1.43}} & \multirow{2}{*}{\underline{1.33}} & \multirow{2}{*}{\underline{2.55}} & \multirow{2}{*}{\underline{1.53}} & \multirow{2}{*}{\underline{0.10}} \\
& (Ours) &&&&&&& \\ \bottomrule[1.5pt]
\end{tabular}
}
\label{tab:kitti}
\end{table}

\subsection{Comparison with State-of-the-Art Methods}
We evaluate our method against state-of-the-art methods on the TartanAir, SceneFlow, and KITTI2015 datasets.
For a fair comparison, video-based baselines are restricted to online methods, excluding offline ones that exploit future frames; representative per-frame methods are also reported for reference.
We evaluate our method on the TartanAir using the train-test split provided by TemporalStereo to compare against other online video methods reported in \cref{tab:tartan}.
Our approach exhibits its most significant advantage in occluded regions, decisively outperforming online video methods that falter in these areas due to their reliance on unreliable temporal warping.
This highlights the robustness of our method, which effectively handles ill-posed areas, even in the presence of misleading or missing temporal priors.

The results on SceneFlow in \cref{tab:sceneflow} demonstrate that our method achieves state-of-the-art performance among online video stereo approaches, attaining an EPE of 0.45 and outperforming TC-Stereo by 10$\%$. Qualitative comparisons in \cref{fig:qual} further highlight the advantages of our approach: it generates temporally consistent depth maps across video frames, whereas conventional methods such as TC-Stereo and IGEV-Stereo struggle to maintain consistency.

To evaluate performance under real-world conditions, we assess DEQ-Stereo on the KITTI2015 dataset, as shown in \cref{tab:kitti}.
Without employing additional modules for high-frequency detail preservation, our method achieves higher accuracy than most existing approaches~\cite{wang2024selective, chen2024mocha, zeng2024temporally} in background and overall metrics, while showing a moderate decline in the foreground region.
In the per-frame mode, DEQ-Stereo maintains strong accuracy and operates 3.12$\times$ faster than MoCha-Stereo~\cite{chen2024mocha}, demonstrating an improved balance between accuracy and efficiency.
In the video mode, our method achieves accuracy comparable to TC-Stereo in background regions, with only minor degradation in foreground and overall metrics.
This performance is obtained without auxiliary camera parameters or complex temporal warping, while running 10$\%$ faster, underscoring the effectiveness of our temporal modeling within the DEQ framework.

\begin{figure*}[t!]
     \centering
     \includegraphics[width=0.9 \linewidth]{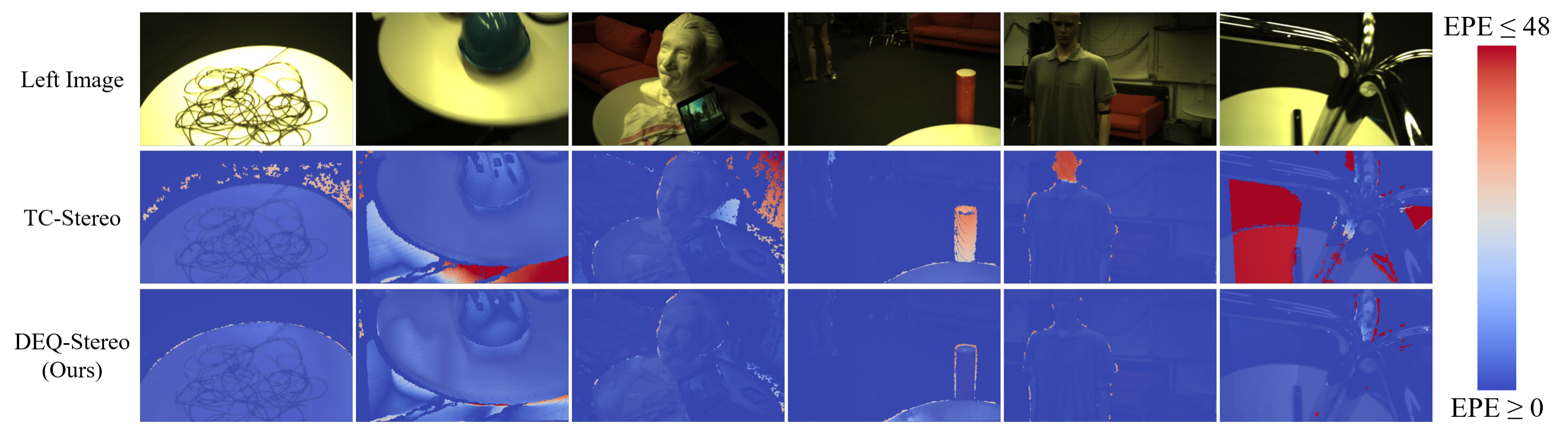}
     \caption{\textbf{Visualization of disparity results on ETH3D SLAM dataset}. DEQ-Stereo produces accurate and stable disparities across diverse real-world scenes, demonstrating robustness to photometric variations and challenging scene structures.}
     \label{fig:eth3d_slam}
\end{figure*}

\begin{table}[t!]
    \centering
    \begin{minipage}[t]{0.48\textwidth}
        \centering
        \caption{Generalization Performance on ETH3D SLAM.}
        \label{tab:eth3d_slam}
        \resizebox{\linewidth}{!}{
        \begin{tabular}{c|ccc}
            \toprule[1.5pt]
            Method              & EPE (px)          & Bad 3.0 (\%)      & Bad 1.0 (\%)       \\ \midrule
            CODD                & 7.78              & 53.02             & 33.19              \\
            TC-Stereo           & 7.17              & \textbf{30.63}    & \textbf{14.70}     \\ \midrule
            DEQ-Stereo (Ours)   & \textbf{3.18}     & 34.85             & 14.80              \\
            \bottomrule[1.5pt]
        \end{tabular}
        }
    \end{minipage}
    \hfill 
    \begin{minipage}[t]{0.48\textwidth}
        \centering
        \caption{Generalization Performance on Spring.}
        \label{tab:spring}
        \resizebox{\linewidth}{!}{
        \begin{tabular}{c|ccc}
            \toprule[1.5pt]
            Method                    & EPE (px)          & Bad 3.0 (\%)      & Bad 1.0 (\%)       \\ \midrule
            TemporalStereo            & 31.36             & 31.40             & 38.20              \\
            CODD                      & \underline{6.82}  & 8.18              & \underline{14.70}  \\
            TC-Stereo                 & 9.91              & \textbf{5.45}     & \textbf{11.92}     \\ \midrule
            DEQ-Stereo (Ours)         & \textbf{2.78}     & \underline{5.74}  & 15.20              \\
            \bottomrule[1.5pt]
        \end{tabular}
        }
    \end{minipage}
\end{table}

\subsection{Zero-Shot Generalization to Real-World}
We evaluate the real-world generalization of our method by testing a TartanAir-pretrained model on the ETH3D SLAM dataset, as presented in \cref{tab:eth3d_slam} and \cref{fig:eth3d_slam}.
Compared to TC-Stereo, DEQ-Stereo achieves a substantially lower EPE, with a 2.25$\times$ improvement, indicating that large disparity errors are significantly reduced.
While TC-Stereo reports slightly lower Bad 3.0 values in some frames, these metrics are sensitive to the distribution of small errors and do not fully capture the presence of large outliers.
In contrast, the consistently low EPE of DEQ-Stereo demonstrates that it produces more reliable disparity estimates across all valid pixels, reducing the likelihood of severe mismatches near challenging regions such as object boundaries and textureless areas.
This analysis confirms that our method not only generalizes well to unseen real-world scenes but also maintains stability and robustness in difficult stereo conditions.

\subsection{Zero-Shot Generalization to Unseen Domains}
We evaluate the generalization capability of models pretrained on SceneFlow~\cite{sceneflow} on the high-resolution (1920$\times$1080) Spring~\cite{mehl2023spring} dataset (\cref{tab:spring}). While many online methods sacrifice generalization performance by reducing feature channels or limiting iterations to ensure real-time inference, our method maintains the same feature capacity as IGEV-Stereo~\cite{xu2023iterative} while utilizing the DEQ framework to eliminate redundant computation.
Consequently, our model handles complex scene dynamics more effectively than alternative architectures. For instance, TemporalStereo~\cite{zhang2023temporalstereo} struggles on Spring because its 3D CNN-based aggregation fails to capture global context at high resolutions, whereas our recurrent-unit approach naturally expands the receptive field via iteration. As a result, our method achieves a 3.56$\times$ improvement in EPE over TC-Stereo~\cite{zeng2024temporally} and a 2.45$\times$ improvement over the second-best method, CODD~\cite{li2023temporally}. While Bad 3.0 and Bad 1.0 metrics remain competitive, this substantial EPE drop indicates that our approach significantly minimizes large disparity errors.

\begin{table}[t!]
    \centering
    \begin{minipage}[t]{0.48\textwidth}
        \centering
        \caption{Ablation study for DEQ-Stereo on TartanAir.}
        \label{tab:module_ablation}
        \resizebox{\linewidth}{!}{
        \begin{tabular}{c|ccc|ccc}
            \toprule[1.5pt]
            \multirow{2}{*}{Method}    & DEQ         & \multirow{2}{*}{ConvSGU}   & \multirow{2}{*}{TST}  & EPE   & Bad 3.0   & Time   \\
                                       & solver      &                            &                       & (px)  & (\%)      & (s)    \\ \midrule
            Baseline                   &             &                            &                       & 0.40 & 1.95 & 0.253 \\ \midrule
            w/o DEQ                    &             & \checkmark                 & \checkmark            & 0.45 & 2.22 & 0.246 \\
            A                          & \checkmark  &                            &                       & 0.41 & 2.04 & 0.287 \\
            B                          & \checkmark  &                            & \checkmark            & 0.49 & 2.39 & 0.192 \\
            C                          & \checkmark  & \checkmark                 &                       & \textbf{0.38}  & \textbf{1.87} & \underline{0.190} \\ \midrule
            Full Model                 & \checkmark  & \checkmark                 & \checkmark            & \underline{0.39}  & \underline{1.88} & \textbf{0.125} \\
            \bottomrule[1.5pt]
        \end{tabular}
        }
    \end{minipage}
    \hfill 
    \begin{minipage}[t]{0.48\textwidth}
        \centering
        \caption{Comparison with foundation model on KITTI2015. A method$^{\dag}$ is fine-tuned on the KITTI Eigen split, while others are trained only on the SceneFlow dataset.}
        \label{tab:foundation}
        \resizebox{\linewidth}{!}{
        \begin{tabular}{c|ccc|c|cc}
            \toprule[1.5pt]
            \multirow{2}{*}{Method}   & \multicolumn{3}{c|}{Bad 1.0} & EPE & Time & Params \\ \cmidrule{2-4}
                                      & NOC     & OCC     & ALL     & (px) & (s)      & (M) \\ \midrule
            StereoAnywhere            & 22.00   & 59.12   & 22.63   & 0.97 & 1.619 & 346.75 \\
            MonSter                   & 19.83   & 46.34   & 20.30   & 0.88 & 0.673 & 388.69 \\
            DEFOM-Stereo              & 23.82   & 57.05   & 24.39   & 1.06 & 0.530 & 382.62 \\
            DEQ-Stereo (Ours)         & 25.66   & 52.78   & 26.15   & 1.33 & 0.216 & 12.60  \\ \midrule
            DEQ-Stereo$^{\dag}$ (Ours) & \textbf{2.57}    & \textbf{12.88}   & \textbf{2.73}    & \textbf{0.27} & \textbf{0.168} & \textbf{12.60}  \\
            \bottomrule[1.5pt]
        \end{tabular}
        }
    \end{minipage}
\end{table}

\subsection{Ablation study for DEQ-Stereo}

\begin{figure}[t!]
    \centering
    \begin{minipage}[t]{0.45\textwidth}
        \centering
        \includegraphics[width=\linewidth]{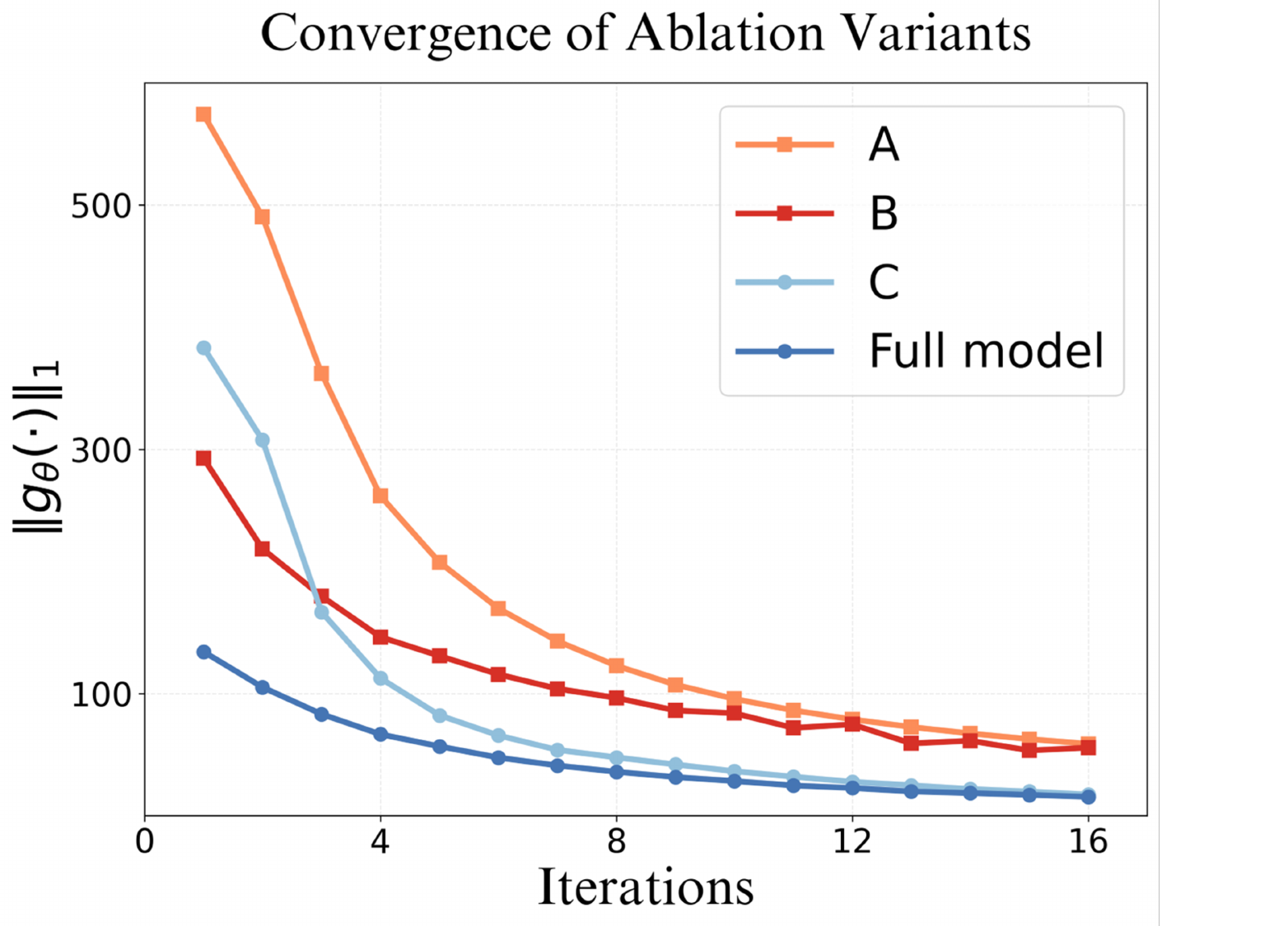}
        \caption{\textbf{Convergence comparison across ablation variants}. This graph shows the maximum norm of $g_\theta(\cdot)$ across DEQ solver iterations on TartanAir. Labels correspond to those in \cref{tab:module_ablation}.} 
        \label{fig:ablation}
    \end{minipage}
    \hfill 
    \begin{minipage}[t]{0.45\textwidth}
        \centering
        \includegraphics[width=\linewidth]{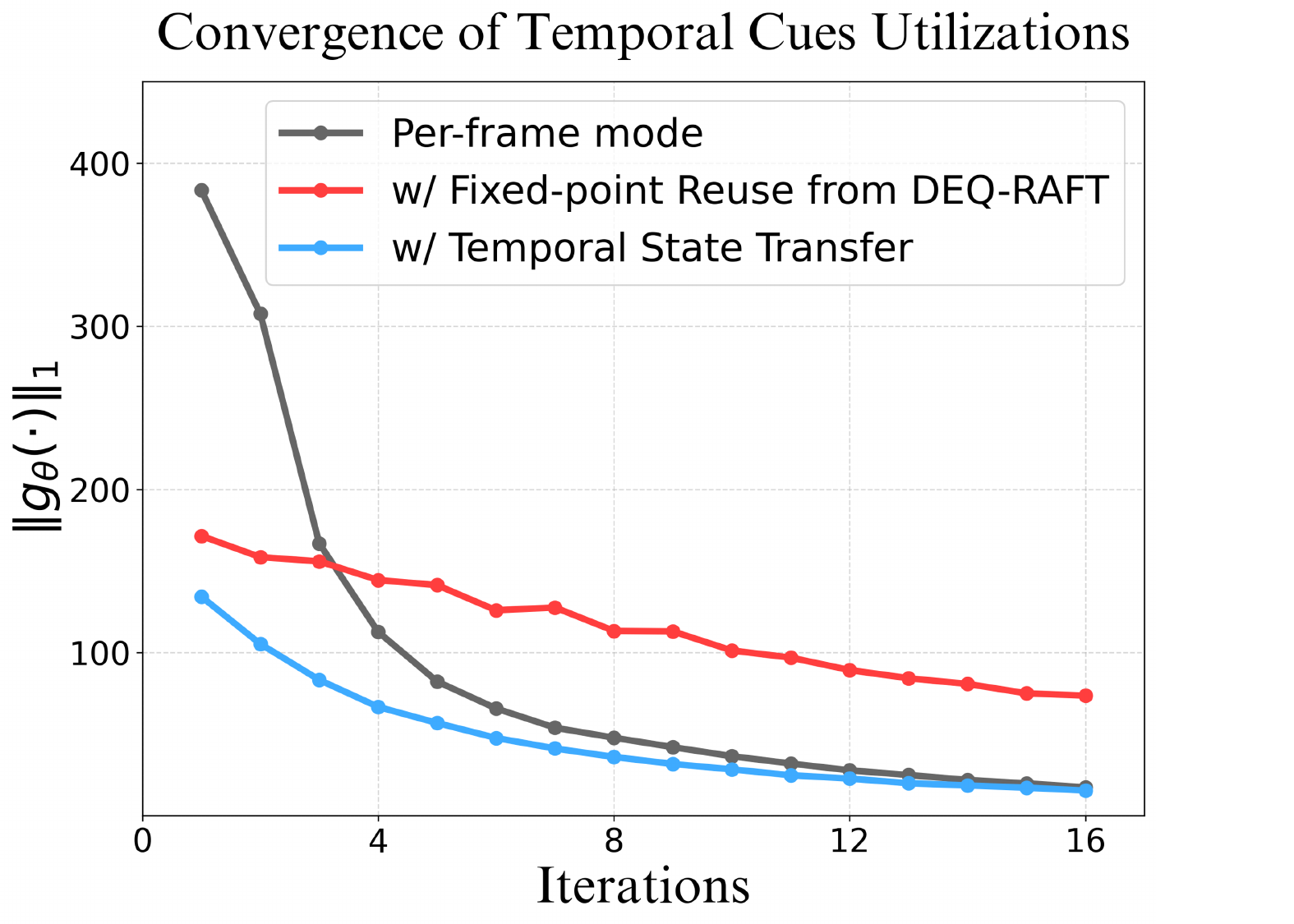}
        \caption{\textbf{Comparison of Methods for Using Temporal Information}. This compares the convergence of Fixed-point Reuse from DEQ-Flow with our TST on TartanAir.} 
        \label{fig:convergence}
    \end{minipage}
\end{figure}

\noindent \textbf{Deep Equilibrium Solver.}
\cref{tab:module_ablation} summarizes our ablation study conducted on the TartanAir dataset. A naive DEQ plug-in leads to degraded accuracy and increased runtime. As illustrated in \cref{fig:ablation}, this failure arises from the ConvGRU’s unstable convergence, which struggles to reach a fixed point under the complex stereo dynamics. This behavior resembles the convergence issues reported in DEQ-Flow~\cite{bai2022deep}, underscoring the need for mechanisms that ensure faster and more stable equilibrium.
Comparing our full model with a recurrent variant without the DEQ solver further supports this: removing the solver increases the EPE by 15.4\% (0.39 $\rightarrow$ 0.45), while our full model achieves higher accuracy at nearly twice the speed. The recurrent variant suffers from initialization sensitivity due to reduced non-linearity and direct reuse of hidden states in TST. In contrast, our DEQ formulation treats inference as an optimization process that iteratively refines the hidden state toward a stable equilibrium, mitigating initialization issues and improving both accuracy and efficiency.

\noindent \textbf{Convolutional Simplified Gated Unit.}
Replacing ConvGRU with our simplified ConvSGU results in noticeably different convergence behavior. As shown in \cref{fig:ablation}, this stems from reduced non-linearity, which enables smoother dynamics and convergence to a deeper equilibrium state where the residual function's norm is closer to zero. This deeper convergence is mathematically akin to performing more effective refinement iterations, explaining the higher accuracy and significantly faster inference reported in \cref{tab:module_ablation}. Further analysis of gating mechanisms and their design choices is included in the supplementary material.

\noindent \textbf{Temporal State Transfer.}
Initializing the solver with the previous frame’s equilibrium state leads to a substantial acceleration in inference, reducing runtime by 34\% in \cref{tab:module_ablation}, while incurring only a minor drop in accuracy, around 0.5\% in Bad 3.0, under challenging conditions. This trade-off makes it highly suitable for real-time video applications.

\subsection{Comparison with Foundation Models}
Recent works~\cite{bartolomei2025stereo, wen2025foundationstereo, jiang2025defom, cheng2025monster} have actively explored integrating the complementary strengths of stereo matching and monocular depth estimation by leveraging foundation depth estimation networks~\cite{yang2024depth}.
To fairly compare with these foundation-based approaches, we evaluate DEQ-Stereo on KITTI2015 dataset and analyze its practicality and efficiency.
All models are trained on SceneFlow and evaluated on KITTI2015 in a zero-shot manner, except for DEQ-Stereo$^{\dag}$, which is fine-tuned on KITTI Eigen split.
As shown in \cref{tab:foundation}, while the foundation-based methods demonstrate strong cross-domain generalization, they require substantially higher computation.
In contrast, DEQ-Stereo attains competitive zero-shot accuracy while running 2.5--7.5$\times$ faster with 27--31$\times$ fewer parameters, highlighting its superior efficiency.
Furthermore, when fine-tuned (DEQ-Stereo$^{\dag}$), both accuracy and inference speed further improve, indicating that our adaptive equilibrium formulation effectively benefits from task-specific adaptation.

\subsection{Comparative Analysis with DEQ-Flow}

To validate that our TST better promotes convergence under the DEQ paradigm compared to the Fixed-point Reuse proposed in DEQ-Flow, we compare their convergence behaviors across iterations in \cref{tab:raft_iters} and \cref{fig:convergence}.
While Fixed-point Reuse provides strong temporal cues, propagating a final prediction with its inherent sharp discontinuities creates a rigid prior.
This prior can become temporally misaligned in dynamic scenes, leading to unstable convergence.
As shown in \cref{tab:raft_iters}, this instability results in a notable degradation in performance compared to the case without temporal cues, indicating that Fixed-point Reuse can hinder convergence rather than accelerate it.
In contrast, TST achieves more stable and faster convergence by propagating only the converged hidden state $\mathbf{h}^*$.
This enables the model to reach deeper fixed points more efficiently, maintaining consistent accuracy across iterations without performance degradation.
By avoiding the reuse of potentially unreliable explicit predictions and instead leveraging an abstract geometric context as a warm-start, TST enhances both robustness and efficiency within the DEQ framework.

\begin{table}[t!]
	\centering
	\caption{Comparison of TST and Fixed-point Reuse on TartanAir.}
        \resizebox{0.8\textwidth}{!}{
        \setlength{\tabcolsep}{12pt}
        \begin{tabular}{c|cccccc}
		\toprule[1.5pt]
		Iterations            & 1     & 2     & 4     & 8     & 16     & Adaptive            \\ \midrule
		w/o Temporal Cue      & 0.64  & 0.53  & 0.47  & 0.42  & \textbf{0.39} & \textbf{0.39} \\
		Fixed-point Reuse     & 13.39 & 9.18  & 6.79  & 5.77  & 4.94 & 6.70 \\
		TST                   & 0.59  & 0.52  & 0.47  & 0.42  & 0.40 & \textbf{0.39} \\
		\bottomrule[1.5pt]
	\end{tabular}
        }
	\label{tab:raft_iters}
\end{table}
\section{Conclusion}
We introduce DEQ-Stereo, an efficient iterative stereo matching framework rooted in the Deep Equilibrium (DEQ) paradigm. By replacing traditional fixed-iteration update loops with an implicit fixed-point formulation, our architecture adaptively balances accuracy and computational efficiency. Specifically, the proposed ConvSGU accelerates and stabilizes per-frame convergence by eliminating unnecessary non-linearities, while the TST mechanism enables effective cross-frame warm-starting. Together, these components allow DEQ-Stereo to achieve state-of-the-art or highly competitive performance across multiple challenging benchmarks, demonstrating strong robustness in occluded and dynamic regions without relying on explicit, computationally expensive temporal warping.

\subsection*{Acknowledgments}
This work was supported by Korea Research Institute for defense Technology planning and advancement through Defense Innovation Vanguard Enterprise Project, funded by Defense Acquisition Program Administration (R230206), the National Research Foundation of Korea(NRF) grant funded by the Korea government(MSIT) (RS-2026-25480865), LG AI STAR Talent Development Program for Leading Large-Scale Generative AI Models in the Physical AI Domain (RS-2025-25442149) and the Advanced GPU Utilization Support Program funded by the Government of the Republic of Korea (Ministry of Science and ICT) (02-26-01-0057).

%
%
\bibliographystyle{splncs04}
\bibliography{main}


\end{document}


\title{DEQ-Stereo: Deep Equilibrium Video Stereo Matching via Adaptive Convergence}

\titlerunning{DEQ-Stereo}

\author{Kiljoon Han\inst{1}\orcidlink{0009-0001-7276-8858} \and
Kyumin Hwang\inst{2}\orcidlink{0009-0000-7878-697X} \and
Wonhyeok Choi\inst{1}\orcidlink{0009-0005-3270-8083} \and
Jeonghoon Kim\inst{1}\orcidlink{0000-0002-7568-4115} \and
Sunghoon Im\inst{2}$^{,\mathsf{\dagger}}$\orcidlink{0000-0001-9776-8101}}

\authorrunning{K.~Han et al.}

\institute{Daegu Gyeongbuk Institute of Science and Technology, Daegu, South Korea \and
Korea Advanced Institute of Science and Technology, Daejeon, South Korea}

\maketitle

\section*{Additional Implementation Details}
\noindent \textbf{ETH3D SLAM Evaluation Setup.}
The ETH3D SLAM~\cite{schops2019bad} dataset provides unrectified stereo pairs together with camera intrinsics and poses for most frames; only a small fraction of frames lack pose annotations. 
For evaluation, we apply stereo rectification to all stereo pairs using the provided calibration parameters. 
Since our method operates on consecutive frame pairs but does not require pose information, we simply restrict the evaluation to frame pairs where ground-truth poses are available for both frames so that all competing methods can be evaluated on the same set of samples. 
Unlike TC-Stereo~\cite{zeng2024temporally}, which removes pixels with EPE $>50$ to mitigate the effect of noisy depth near object boundaries, we retain all valid disparity pixels within the range $0 < \text{disp} < 192$. 
This choice preserves large-error regions and provides a more comprehensive assessment of real-world generalization performance.

\section*{Joint Equilibrium Formulation}
In our Deep Equilibrium framework, we define the joint state as the concatenation of the hidden state and the disparity, denoted as $\mathbf{s} = [\mathbf{h}, \mathbf{d}]$.
Based on this definition, the residual function $g_\theta(\mathbf{s}^k, \mathbf{C})$ at iteration $k$ measures the difference between the current joint state and the output of the update function $f_\theta$.
This residual partitions into two components corresponding to the hidden state and the disparity:
\begin{equation}
\label{equ:joint}
    \begin{aligned}
        g_\theta(\mathbf{s}^k, \mathbf{C}) &= \mathbf{s}^k - f_\theta(\mathbf{s}^k, \mathbf{C}) 
        \\&= [\mathbf{h}^k - \mathbf{h}^{k+1}, \mathbf{d}^k-\mathbf{d}^{k+1}] = [g^k_{\mathbf{h}}, g^k_{\mathbf{d}}],
    \end{aligned}
\end{equation}
where $g^k_{\mathbf{h}}$ and $g^k_{\mathbf{d}}$ represent the individual residuals for the hidden state and the disparity at iteration $k$, respectively.

To find the equilibrium state where $g_\theta(\mathbf{s}^k, \mathbf{C}) \rightarrow 0$, the numerical solver approximates the inverse Jacobian matrix $\mathbf{B}^{k}_{g_\theta}$ using the Broyden method.
Because the state vector $\mathbf{s}$ encapsulates two distinct variables, this approximated Jacobian operates as a block matrix.
This structure explicitly captures the mutual dependencies between the hidden state and the disparity during the update process:
\begin{equation}
\label{equ:joint}
    \begin{gathered}
        \mathbf{B}^{k}_{g_\theta} \!=\! \begin{bmatrix} \mathbf{B}_{\mathbf{hh}}^k \! & \! \mathbf{B}_{\mathbf{hd}}^k \\ \mathbf{B}_{\mathbf{dh}}^k \! &\!  \mathbf{B}_{\mathbf{dd}}^k \end{bmatrix}, \\
        \mathbf{s}^{k+1} \!=\! \mathbf{s}^{k} \!-\! \alpha\mathbf{B}^{k}_{g_\theta}g_\theta(\mathbf{s}^{k})\! =\! \begin{bmatrix}\mathbf{h}^k - \alpha \left( \mathbf{B}_{\mathbf{hh}}^k g_{\mathbf{h}}^k + \mathbf{B}_{\mathbf{hd}}^k g_{\mathbf{d}}^k \right) \\ \mathbf{d}^k - \alpha \left( \mathbf{B}_{\mathbf{dh}}^k g_{\mathbf{h}}^k + \mathbf{B}_{\mathbf{dd}}^k g_{\mathbf{d}}^k \right) \end{bmatrix},
        \\
        \mathbf{B}^{k}_{g_\theta} = \mathbf{B}^{k-1}_{g_\theta} + \frac{\Delta \mathbf{s}^{k} - \mathbf{B}^{k-1}_{g_\theta}\Delta g^{k}_\theta}{\left(\Delta \mathbf{s}^{k}\right)^\top \mathbf{B}^{k-1}_{g_\theta}\Delta g^{k}_\theta} \left(\Delta \mathbf{s}^{k}\right)^\top \mathbf{B}^{k-1}_{g_\theta},
    \end{gathered}
\end{equation}
where the off-diagonal blocks $\mathbf{B}_{\mathbf{hd}}^k$ and $\mathbf{B}_{\mathbf{dh}}^k$ capture the cross-variable interactions, allowing the hidden state residual $g^k_{\mathbf{h}}$ to directly influence the disparity update, and vice versa. The matrix $\mathbf{B}^{k}_{g_\theta}$ is dynamically updated at each step via the Broyden rank-one rule with $\Delta \mathbf{s}^{k} = \mathbf{s}^{k} - \mathbf{s}^{k-1}$ and $\Delta g^{k}_\theta = g_\theta(\mathbf{s}^{k}) - g_\theta(\mathbf{s}^{k-1})$.
By explicitly modeling these mutual dependencies, the solver actively compensates for how a change in the feature space ($\mathbf{h}$) impacts the geometric prediction ($\mathbf{d}$). This coupled optimization steers both variables to converge simultaneously, preventing unstable oscillations and ensuring the system reaches a stable joint equilibrium where both residuals vanish together.

\begin{figure*}[t!]
    \centering
    \begin{minipage}[t]{0.480\linewidth}
        \centering
	\resizebox{0.95\textwidth}{!}{
        \begin{tabular}{c|cccccc}
		\toprule[1.5pt]
		Threshold $\epsilon$    & 50     & 30    & 10    & 1     & $1e\text{-}3$ \\ \midrule
		EPE (px)                & 0.47   & 0.45  & 0.45  & 0.45  & 0.45      \\
		Inference Time (s)      & 0.17   & 0.22  & 0.33  & 0.40  & 0.40      \\
		\bottomrule[1.5pt]
	\end{tabular}
        }

        \captionof{table}{Ablation study for Threshold $\epsilon$ on Sceneflow.}
        \label{tab:threshold_ablation}
    \end{minipage}
    \hfill
    \begin{minipage}[t]{0.480\linewidth}
        \centering
        
\resizebox{0.95\textwidth}{!}{
\begin{tabular}{c|ccc|cc}
\toprule[1.5pt]
Model      & Reset      & Update      & Forget      & EPE (px)      & Bad 3.0 (\%) \\ \midrule
Baseline   &            &             &             & 0.67          & 3.41         \\ \midrule
A          & \checkmark &             &             & 0.70           & 3.54         \\
B          &            & \checkmark  &             & 0.49          & 2.59         \\
ConvGRU    & \checkmark & \checkmark  &             & 0.47          & 2.48         \\ \midrule
ConvSGU    &            & \checkmark  & \checkmark  & \textbf{0.45} & \textbf{2.37} \\
\bottomrule[1.5pt]
\end{tabular}
}
        \captionof{table}{Ablation study on gate of recurrent on Sceneflow.}
        \label{tab:gate_ablation}
    \end{minipage}
\end{figure*}

\section*{Ablation study for Threshold}
During training, we set the convergence threshold $\epsilon$ to $1e\text{-}3$ to ensure stable optimization, whereas during inference, we relax it to 30 to improve computational efficiency.
To analyze the effect of $\epsilon$, we evaluate accuracy and inference time under different threshold values, as summarized in \cref{tab:threshold_ablation}.  
When $\epsilon=50$, the model terminates too early, before reaching a sufficiently stable fixed point, which results in a slightly higher EPE of 0.47 compared to 0.45 for $\epsilon \leq 30$. This indicates that the model has not fully converged, leading to occasional larger disparity errors despite the shortest inference time.  
Thresholds of $\epsilon=30$ and below allow the model to reach a sufficiently converged solution, yielding consistent accuracy (EPE = 0.45).  
However, lowering the threshold further to 10, 1, or $1e\text{-}3$ causes additional iterations to occur even after the hidden states have essentially converged. While these extra iterations do not improve EPE, they increase inference latency substantially, representing a 1.81$\times$ increase in runtime without any accuracy benefit.
These observations suggest that setting $\epsilon=30$ during inference achieves an optimal balance between accuracy and efficiency, maintaining a fully converged solution with minimal computational overhead.

\section*{Ablation study for Recurrent Unit}
To assess the suitability of ConvSGU for the DEQ paradigm, we conduct ablation studies on the gating mechanisms within the recurrent unit on the SceneFlow~\cite{sceneflow} dataset, as shown in \cref{tab:gate_ablation}.
Using only the reset gate (\cref{tab:gate_ablation} - A) results in lower performance compared to the baseline, indicating that restricting the influence of past states alone is insufficient.
Introducing the update gate (\cref{tab:gate_ablation} - B) significantly improves performance by enabling long-term information flow.
Further improvements are observed when the reset gate is added (\cref{tab:gate_ablation} - ConvGRU), as it helps refine the candidate state using previous context.
However, the multiplicative interaction introduced by the reset gate increases non-linearity and can destabilize convergence within the DEQ solver.
To address this, we remove the reset gate and instead introduce a forget gate (\cref{tab:gate_ablation} - ConvSGU), which acts as a dynamic damping factor to control residual flow between iterations.
This modification results in smoother dynamics and achieves the best performance across both EPE and Bad 3.0.
Inspired by the additive transition mechanism of the Recurrent Additive Network (RAN)~\cite{lee2017recurrentadditivenetworks}, ConvSGU follows a similar additive update principle but extends it to a convolutional form that maintains spatial expressiveness and stability within the DEQ framework.

\begin{figure}[t!]
     \centering
     \includegraphics[width=0.99\linewidth]{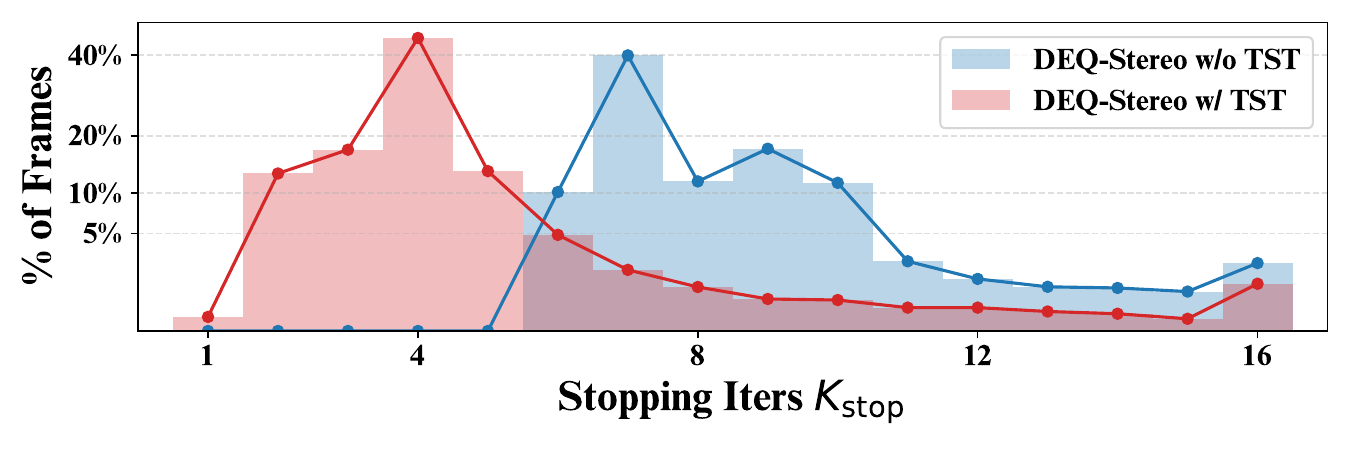}
     \caption{\textbf{Stopping iteration $K_{\text{stop}}$ distribution.} Comparison with and without TST on the TartanAir test set. The y-axis indicates the percentage of frames converging at a given iteration.}
     \label{fig:stop}
\end{figure}

\section*{Analysis of Stopping Iteration $K_{\text{stop}}$}
\cref{fig:stop} illustrates the distribution of stopping iterations $K_{\text{stop}}$ to provide a deeper understanding of how Temporal State Transfer (TST) accelerates inference. Without TST, the DEQ-Stereo requires a relatively high number of iterations to reach equilibrium. The distribution for this setting peaks at 7 iterations, accounting for approximately 40\% of the frames.

When TST is enabled, this temporal warm-starting induces a substantial leftward shift in the distribution. The peak moves down to 4 iterations, where over 40\% of the frames successfully converge, and the proportion of frames requiring 8 or more iterations drops significantly. By selectively transferring the converged hidden state $\mathbf{h}_{t-1}^*$ from the previous frame, the solver begins its updates much closer to the true joint equilibrium state of the current frame. Because consecutive video frames exhibit strong temporal coherence, this informed initialization drastically reduces the required number of iterations, thereby improving inference efficiency while retaining the implicit refinement guarantees.

\begin{table}[t]
    \centering
    \setlength{\tabcolsep}{5pt}
    \caption{\textbf{Convergence under Varying Motion.} $\tau$ denotes the frame interval simulating varying inter-frame motion.}
    \resizebox{0.85\textwidth}{!}{
    \setlength{\tabcolsep}{12pt}
	\begin{tabular}{c|c|ccccc}
		\toprule[1.5pt]
        Method                 & w/o TST           & \multicolumn{5}{c}{w/ TST} \\ \midrule
		Temporal Stride $\tau$ & 1                 & 1                 & 2                 & 4         & 8         & 16         \\ \midrule
		EPE                    & 0.39              & 0.39              & 0.39              & 0.39      & 0.39      & 0.40      \\
        Avg. $K_{\text{stop}}$ & 8.29              & \textbf{4.17}     & \underline{4.62}  & 5.15      & 5.70      & 6.29      \\
		\bottomrule[1.5pt]
	\end{tabular}
    }
    \label{tab:time_stride}
\end{table}
\section*{Robustness to Large motion.}
We further analyze the robustness of our method across varying temporal intervals, as detailed in \cref{tab:time_stride}.
As the time stride increases, inter-frame motions become larger, making the temporal prior $\mathbf{h}_{t-1}^*$ less aligned with the optimal state of the current frame.
Methods relying on explicit temporal feature warping typically suffer severe performance degradation under these conditions due to warping failures.
Our TST mechanism, however, uses the previous state merely as an informed initialization rather than a hard constraint.
When faced with a larger time stride, the DEQ solver dynamically compensates by executing additional iterations to securely reach the new joint equilibrium.
This adaptive convergence ensures that spatial accuracy degrades gracefully, demonstrating the robustness of our implicit formulation against low frame rates or fast-moving scenarios.

\section*{Temporal Consistency Analysis}
\cref{tab:eth3d_tepe} compares the Temporal End-Point Error ($\text{TEPE}=|\Delta \mathbf{d} - \Delta \mathbf{d}_{\text{GT}}|$)~\cite{karaev2023dynamicstereo, li2023temporally}.
While our method shows a slightly higher TEPE than TC-Stereo, it achieves a significantly lower EPE.
This discrepancy arises because TEPE measures consistency by comparing adjacent frames, inherently penalizing jitter.
Consequently, a model that maintains a persistent error yields a better TEPE than a model that actively corrects it, failing to properly capture error accumulation.

    \begin{table*}[t]
    \caption{Quantitative results on the ETH3D SLAM with temporal metrics.}
    \centering
    \resizebox{0.85\textwidth}{!}{
    \setlength{\tabcolsep}{12pt}
	\begin{tabular}{c|cc|cc}
		\toprule[1.5pt]
		Method              & EPE (px)          & Bad 3.0 (\%)      & TEPE (px)     & T-Bad 3.0 (\%) \\ \midrule
		CODD                & 9.43              & 29.65             & 3.44          & 18.89         \\
		TC-Stereo           & \underline{7.64}  & \underline{15.04} & \textbf{1.70} & \textbf{7.84} \\ \midrule
		DEQ-Stereo (Ours)   & \textbf{3.18}     & \textbf{14.80}    & \underline{1.95} & \underline{10.25}\\
		\bottomrule[1.5pt]
	\end{tabular}
    }
    \label{tab:eth3d_tepe}
    \end{table*}

\begin{figure*}[t!]
     \centering
     \includegraphics[width=1.0 \linewidth]{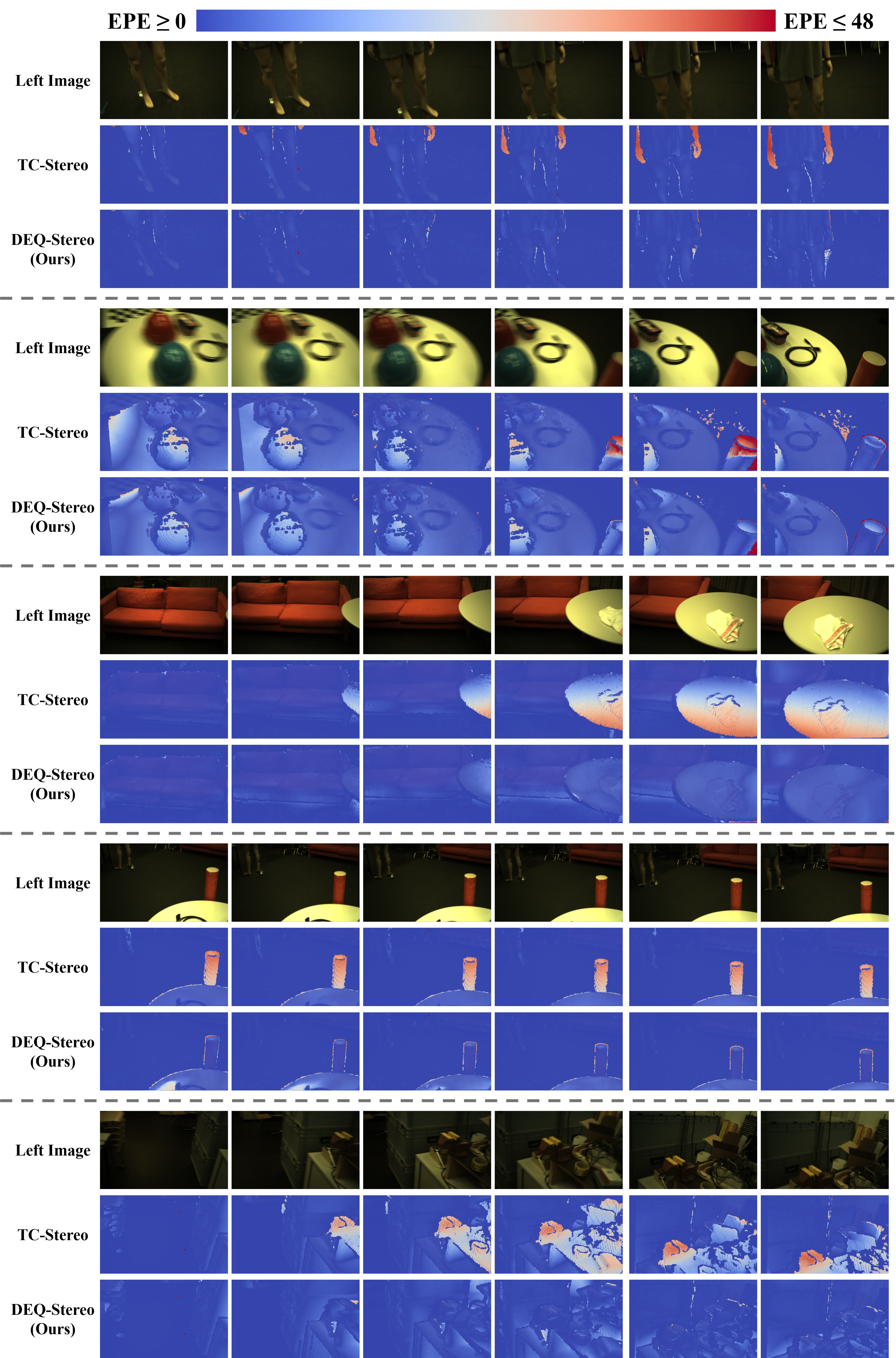}
     \caption{\textbf{Additional qualitative results on ETH3D SLAM dataset}. The video sequences progress from left to right.}
     \label{fig:eth3d_slam_video}
\end{figure*}

\begin{figure*}[t!]
     \centering
     \includegraphics[width=1.0 \linewidth]{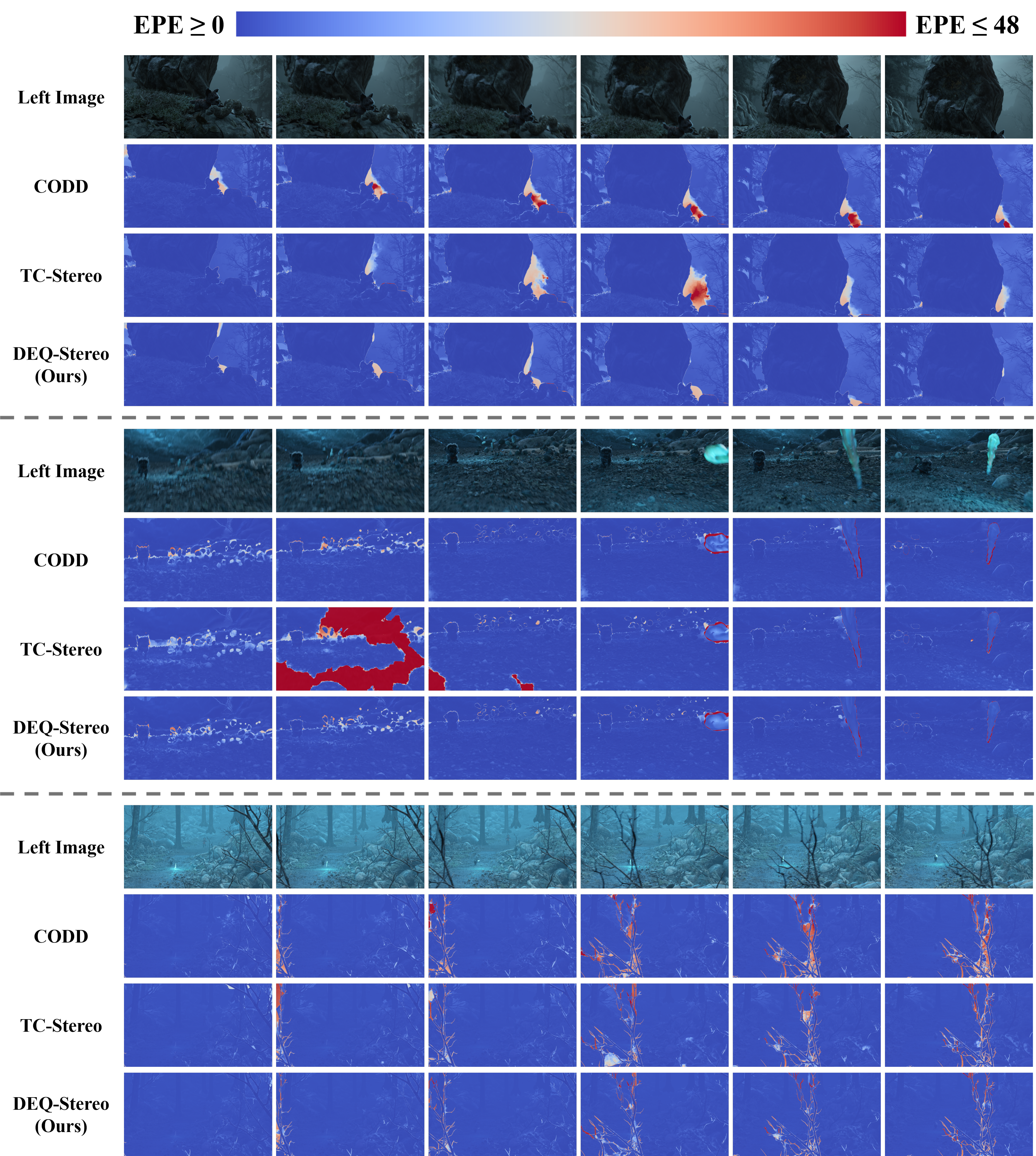}
     \caption{\textbf{Additional qualitative results on Spring dataset}. The video sequences progress from left to right.}
     \label{fig:spring_video}
\end{figure*}

As illustrated in \cref{fig:eth3d_slam_video} and \cref{fig:spring_video},
TC-Stereo~\cite{zeng2024temporally} and CODD~\cite{li2023temporally}, which rely on explicit temporal warping, propagate large errors frame-to-frame.
In contrast, our approach mitigates this error accumulation.
Rather than explicitly transferring predictions, we utilize the hidden feature of the previous frame as the initial state to find a new convergence point.
This mechanism allows the model to actively recover from previous errors.
While this recovery process can induce minor jitter—explaining the slight drop in TEPE—it prevents long-term error propagation, resulting in temporally consistent depth estimation.

\section*{Limitations}
While our method effectively reduces redundant computation via adaptive termination and demonstrates robust performance, certain limitations remain. Because our framework is primarily designed for cost-efficient compatibility with the DEQ paradigm, it does not explicitly model video-specific challenges such as severe occlusions or radical motion. As a result, the TST mechanism can suffer from minor accuracy degradation under highly dynamic motions. Future work will focus on developing adaptive TST mechanisms to handle rapid temporal variations and exploring lightweight architectures to further optimize the trade-off between model complexity and computational efficiency.

%
%
\bibliographystyle{splncs04}
\bibliography{main}

\clearpage

